%% file: main_RLC.tex
\documentclass[10pt]{article} %

\usepackage[preprint]{rlj} %

\usepackage{amssymb}            %
\usepackage{mathtools}          %
\usepackage{mathrsfs}           %
\usepackage{graphicx}           %
\usepackage{subcaption}         %
\usepackage[space]{grffile}     %
\usepackage{url}                %
\usepackage{lipsum}             %
\usepackage{bm}

\usepackage{amsmath}
\DeclareMathOperator*{\argmax}{argmax}

\title{Seldonian Reinforcement Learning for \\Ad Hoc Teamwork}

\setrunningtitle{Seldonian Reinforcement Learning for Ad Hoc Teamwork}

\author{Edoardo Zorzi\textsuperscript{1,$*$}, Alberto Castellini\textsuperscript{1,$*$}, Leonidas Bakopoulos\textsuperscript{2},\\ Georgios Chalkiadakis\textsuperscript{2}, Alessandro Farinelli\textsuperscript{1}}

\emails{\{name.surname\}@univr.it, \ \{lbakopoulos,gchalkiadakis\}@tuc.gr}

\affiliations{
$^{1}$\textbf{Università degli Studi di Verona, Verona, Italy}\\
$^{2}$\textbf{Technical University of Crete, Crete, Greece}\\

\par %
$^*$Equal contribution
}

\contribution{
    We formalize the problem of offline Seldonian reinforcement learning in the context of Ad Hoc Teamwork, \new{adding the assumption of having a candidate set of policies suitable for improvement and knowledge about the possible types of policies of the teammates}. The goal is to select optimal policies with statistical guarantees with respect to \new{desirable} behaviors.}
    {
    Seldonian policy optimization was first proposed in \cite{thomas_preventing_2019}. An application to RL for diabetes management was introduced in the same work. The Ad Hoc Teamwork problem was introduced in \cite{stone2010adhoc} %
    with the goal of designing agents that can collaborate with new teammates without prior coordination. No other work formally defines the problem of offline Seldonian optimization for Ad Hoc Teamwork.
    }
    
\contribution{
    We provide an offline reinforcement learning algorithm based on Seldonian optimization for Ad Hoc Teamwork. %
    The algorithm can solve multiagent scenarios due to an extended version of Doubly-Robust Importance Sampling \citep{jiang_doubly_2016}, which estimates \new{policy desirability} by explicitly representing teammate policies and the transition model.
    }
    {
    The Seldonian RL algorithm proposed in \cite{thomas_preventing_2019} cannot scale to large multiagent scenarios because of importance sampling inefficiency. We introduce, into the estimation process, knowledge about the Ad Hoc Teamwork setting to improve the algorithm's efficiency and allow it to scale. Other Seldonian algorithms in the literature \citep{satija2021multiobjectivespibb,thomas_hcpi_2015} cannot deal with policy reliability in Ad Hoc Teamwork.
    }
    
\contribution{
    We evaluate our algorithm in three Ad Hoc Teamwork scenarios, showing that it is consistently reliable and efficient compared to standard ML baselines. 
    }
    {
    The evaluated domains are chain world \citep{chalkiadakis_coordination_2003}, extended blackjack \citep{sutton2020rl}, and level-based foraging \citep{papoudakis2021benchmarking}.
    }


\summary{Most offline RL algorithms return optimal policies but do not provide statistical guarantees on \new{desirable} behaviors. This could generate reliability issues in safety-critical applications, such as in some multiagent domains where agents, and possibly humans, need to interact to reach their goals without harming each other. In this work, we propose a novel offline RL approach, inspired by Seldonian optimization, which returns policies with good performance and statistically guaranteed properties with respect to predefined \new{desirable} behaviors. 
In particular, our focus is on Ad Hoc Teamwork settings, where agents must collaborate with new teammates without prior coordination. 
Our method requires only a pre-collected dataset, a set of candidate policies for our agent, and a specification about the possible policies followed by the other players---it does not require further interactions, training, or assumptions on the type and architecture of the policies. 
We test our algorithm in Ad Hoc Teamwork problems and show that it consistently finds reliable policies while improving sample efficiency with respect to standard ML baselines.
}

\input{packages.tex}

\newcommand{\defeq}{%
  \mathrel{\vbox{\offinterlineskip\ialign{%
    \hfil##\hfil\cr
    $\cdot$\cr
    $=$\cr
}}}}

\newcommand{\dr}{{\hat\rho}_{i,j,k}^{\texttt{DR}}}

\usepackage{bbm}
\newcommand{\pp}{\mathbbm{P}} 

\newcommand{\ee}{\mathbbm{E}}

\newcommand{\R}{\mathbb{R}}

\begin{document}

\maketitle  %

\begin{abstract}
    Most offline RL algorithms return optimal policies but do not provide statistical guarantees on \new{desirable} behaviors. This could generate reliability issues in safety-critical applications, such as in some multiagent domains where agents, and possibly humans, need to interact to reach their goals without harming each other. In this work, we propose a novel offline RL approach, inspired by Seldonian optimization, which returns policies with good performance and statistically guaranteed properties with respect to predefined \new{desirable} behaviors. 
    In particular, our focus is on Ad Hoc Teamwork settings, where agents must collaborate with new teammates without prior coordination. 
    Our method requires only a pre-collected dataset, a set of candidate policies for our agent, and a specification about the possible policies followed by the other players---it does not require further interactions, training, or assumptions on the type and architecture of the policies. 
    We test our algorithm in Ad Hoc Teamwork problems and show that it consistently finds reliable policies while improving sample efficiency with respect to standard ML baselines.

\end{abstract}

\section{Introduction}

Consider a warehouse environment where multiple robots collect and deliver packages from the shelves to the loading area. %
In such a situation, coordination among the agents would be fundamental to increase throughput and avoid conflicts, such as deadlocks and crashes. This coordination, however, can be difficult to achieve if we can control only the policy of a single robot, e.g., due to proprietary software or other issues. The difficulty increases when the other agents follow different policies: some of them, for example, might act conservatively, slowing down to avoid crashes, whereas others might move faster and more recklessly to guarantee predefined throughput. 
Intuitively, in this situation, we would like to tailor the agent's policy to the \textit{type} of the other agents (the \textit{teammates}) it faces. For instance, if we know a nearby robot is conservative, we might want our robot to go faster to deliver packages more quickly. By contrast, if the other robot is not conservative, then we might like our robot to be more cautious to reduce the chance of a crash. 

Such non-coordinated environments are dealt with by Ad Hoc Teamwork (AHT) \citep{stone2010adhoc, albrecht_autonomous_2018, mirsky_survey_2022} strategies.
However, current state-of-the-art AHT algorithms only consider agent returns, not explicit constraints, which are better suited to express more complex behaviors we might want from our agent, such as collision avoidance.

In this work, we propose a novel formalization of these types of problems in the offline setting. In particular, we assume having a large dataset of interactions previously collected by our agent using, possibly, suboptimal \textit{behavior policies}, a set of new \textit{candidate policies}, obtained for example from training state-of-the-art (deep) RL algorithms, and a set of \textit{constraint functions}, used to capture the desired behavior of our agent.
The goal is to return the \textit{best reliable} candidate policy, irrespective of what the other agents are doing.
For \textit{best reliable} policy, we mean the policy with the highest return among the candidates that is also guaranteed to satisfy, probabilistically, the given constraints.

Moreover, we propose an algorithm to solve this problem, obtained by integrating Ad Hoc Teamwork strategies \citep{stone2010adhoc, albrecht_autonomous_2018, mirsky_survey_2022} with the Seldonian optimization framework \citep{thomas_preventing_2019}. Seldonian optimization allows us to introduce probabilistic constraints (based on confidence levels) in offline policy optimization, while AHT suggests explicitly representing the other agent's policy types and the environment transition model in the optimization process, improving sample efficiency with respect to state-of-the-art Seldonian approaches \citep{thomas_preventing_2019}.

In summary, this paper provides the following contributions to the state of the art: {\em (i)} we propose a novel problem formalization for offline AHT where the goal is to obtain the \textit{best reliable} policy from a set of candidates, adopting the Seldonian optimization framework; {\em (ii)} we provide a method that is statistically guaranteed to return a reliable solution; and {\em (iii)} we empirically evaluate our approach on increasingly complex environments, showing that it can scale up to hard AHT domains.

\section{Background}\label{sec:background}

\subsection{Markov Decision Process}
A Markov Decision Process (MDP) \citep{puterman2014, sutton2020rl} is a tuple $\langle S,A,T,R,\gamma \rangle$, where $S$ is the set of states, $A$ is the set of actions, $T: S\times A \times S \rightarrow [0,1]$ is the stochastic transition function, $r: S \times A \rightarrow [R_{\min},R_{\max}]$ is the reward function, and $\gamma \in [0,1]$ is the discount factor. A stochastic policy $P: S \times A \to [0,1]$ defines a probability function from states to actions such that any agent following the policy $P$ takes action $a$, in state $s$, with probability $P(a|s)$. Given a policy $P$ and a MDP, we can define for each state $s \in S$ the value function $V^P(s)$, that is, the discounted return that the agent is \textit{expected} to get by following $P$ from $s$ from any timestep $t_0$ onwards: $V^P(s) \defeq \mathbb{E}[\sum_{t=t_0}^{\infty}\gamma^{t-t_0} r(s_t, a_t) | a_t \sim P(\cdot | s_t), s = s_t]$. Likewise, the state-action value function $Q^P(s,a)$ is the expected discounted return that the agent will get by taking action $a$ in state $s$ in the current timestep $t_0$ and following $P$ afterward: $Q^P(s,a) \defeq \mathbb{E}[\sum_{t=t_0}^{\infty}\gamma^{t-t_0}r(s_t, a_t) | a_{t+1} \sim P(\cdot | s_{t+1}), s = s_t, a=a_t]$.

\subsection{Seldonian policy optimization}

In \cite{thomas_preventing_2019} a Seldonian batch approach is proposed. It consists of a new algorithm design framework that shifts focus from maximizing the performance to avoiding undesirable behavior, expressed as a probabilistic constraint. Let $\mathcal{P}$ be a set of stochastic or deterministic policies of interest for an MDP. Let $\mathcal{H}$ be a set of possible histories, where a history $H\in\mathcal{H}$ is a sequence of state-action-reward values collected by interacting with the environment. Each policy, $P \in \mathcal{P}$, induces a distribution over $\mathcal{H}$. We write $H \sim P$ to denote that the history-valued random variable $H$ is generated using the policy $P$. Let $r: \mathcal{H} \rightarrow \R$ be the return function, with $r(H)$ denoting the return of history $H$. The expected return when using solution $P$ can be written as $\mathbb{E}_H[r(H) \ | \ H \sim P ]$.

The Seldonian Optimization Problem for RL is defined as: 
\begin{equation}
\begin{aligned}
&\argmax_{P} \mathbb{E}_H[r(H) \ | \ H \sim P ]\\
&\textrm{s.t.} \ \forall j \in \{1, \ldots, n\} \ Pr(g_j(P) \geq 0) \geq 1-\delta_j \label{eq:Seldonian}
\end{aligned}
\end{equation}
where $g_j(P)$ is a deterministic function that defines a measure of
\new{\emph{desirable behavior}} for policy $P$. This function can be thought of as an `alternative return' one, and can, for example, reward the number of avoided collisions by policy $P$, or reward action coordination in a multiagent environment \citep{marl-book}. \new{Note that, in this work, we consider each $g_j$ as a function measuring the desirability of a policy, and hence the inequality inside $Pr$ contains $\ge$ instead of $\le$, which appears in the original formulation \citep{thomas_preventing_2019}.}

\subsection{Ad Hoc Teamwork}

Ad Hoc Teamwork (AHT) \citep{stone2010adhoc, albrecht_autonomous_2018, mirsky_survey_2022} is defined as the problem of developing agents capable of cooperating on the fly with other unfamiliar agents, without prior coordination. The inputs of the AHT problem are domain knowledge (e.g., an MDP definition of the environment, which expresses both the learner's ability, in terms of actions, and the task, in terms of reward) and a list of teammates with a (possibly incomplete) list of their attributes  (e.g., its possible type and related policy). The output of the problem is the learner, represented by a policy $P$, which might be deterministic or stochastic, static or dynamic, depending on the agent’s sensors, the available communication channels, and the task definition. %
The key AHT assumptions  are \emph{i)} no prior coordination, \emph{ii)} no control over teammates, \emph{iii)} collaborative behaviors of the teammates.

The main subtasks that need to be tackled to solve the AHT problem are \emph{i)} the definition of a knowledge representation, \emph{ii)} modeling teammates behaviors or inferring their types, \emph{iii)} policy generation, \emph{iv)} policy adaptation. A complete review of  
AHT literature %
is available in \cite{mirsky_survey_2022}.

\subsection{Problem definition}

The illustrative case study presented in the introduction is an example of the AHT problem we want to solve. The robot we control (ego-agent) and the other robots (teammates) have no predefined coordination strategy. The goal is to generate an optimal policy for the ego-agent while having only domain knowledge in the form of an MDP definition and the possible types of teammates, which must be inferred. In our problem formulation, we make the additional assumption of having a large dataset of trajectories collected offline by possibly suboptimal behavior policies and a set of candidate policies $\{ P_i\}_{i=1}^\ell$. Furthermore, our goal is to return the best policy (in terms of return) only among those that are \emph{reliable} according to some user-defined constraints, e.g., collision avoidance. In this sense, the problem in the illustrative example is similar to the problem tackled by Seldonian policy optimization.  

The problem we aim to solve in this work is, therefore, integrating AHT \citep{mirsky_survey_2022} with Seldonian policy optimization \citep{thomas_preventing_2019}. The inputs are: the dataset of trajectories $D = \{(H_k, P_k )\}_{k=1}^m$ composed of $m$ histories $H_1, \ldots, H_m$ collected by \textit{known} behavior policies $P_1,\ldots, P_m$, where each history is a sequence of tuples $\langle (s_0,\bm{a}_0, s_1, r_1), (s_1,\bm{a}_1, s_2, r_2),\ldots \rangle$ with $s_t$ the environment state, $\bm{a}_t$ the joint action (of the ego-agent and the teammates), and $r_t$ the ego-agent reward; the reward function $r$ (we assume it is known); a set of candidate policies $\{ P_1, \ldots, P_\ell\}$ for the ego-agent; a set of possible teammate types $\{ \mathcal{T}_1, \ldots, \mathcal{T}_q \}$ with corresponding policies $P_{\mathcal{T}_1}, \ldots, P_{\mathcal{T}_q}$; the number of teammates $p$; and a set of measures of \new{desirable} behavior $g_1, \ldots, g_n$, which we model as function $g_j: S\times A \to \mathbb{R}$, and their required confidence levels $\delta_1, \ldots, \delta_n$. {We define $g_j(P)$ as the expected discounted \new{`return'} $g_j$ over the trajectories obtained by $P$, that is, $g_j(P) \defeq \mathbb{E}_H[g_j(H) | H \sim P]$.}  The output is the best candidate policy $P^\star \in \{ P_1, \ldots, P_\ell\}$, in terms of return $r(P) \defeq \ee_H[r(H) | H \sim P]$, among those that satisfy the probabilistic constraints $g_1, \ldots, g_n$ on \new{desirable} behaviors with a corresponding confidence level.

\section{Method}\label{sec:method}

The main idea behind the proposed approach is to adequately take into account in the performance estimate $\hat{\rho}_{i,j,k}$ of $g_j(P_i)$ ($k\in\{ 1, \ldots,m\}$ is a trajectory) the transition model $T$ and the $p$ teammates, each having an (unknown) type in the set $\mathcal{T}=\{\mathcal{T}_1, \ldots, \mathcal{T}_q\}$ with corresponding policy $P_{\mathcal{T}_1}, \ldots, P_{\mathcal{T}_q}$, and then use a finite sample concentration inequality to satisfy the probability constraint in Eq. \eqref{eq:Seldonian}, while making sure that $\hat{\rho}_{i,j,k}$ is an unbiased estimate of $g_j(P_i)$: $\ee[\hat{\rho}_{i,j,k}] = g_j(P_i)$.

\subsection{Performance estimation with Importance Sampling}

Given a target policy $P_i$, a reward function $r_j$, and a trajectory $H_k$ obtained by a behavior policy $P_k$, an off-policy estimate $\hat{\rho}_{i,j,k}$ of the performance of $P_i$ with respect to the function $g_j$ can be obtained via \textit{Importance Sampling} (IS) \citep{Kahn_1953, doina2000Eligibility}:
\begin{align}&\hat{\rho}_{i,j,k}^{\texttt{IS}} \defeq g_j(H_k) \prod_{(s_t,\mathbf{a}_t)\in H_k} \frac{P_i(\mathbf{a}_t|s_t)}{P_k(\mathbf{a}_t|s_t)}\label{eq:IS}
\end{align}
where $s_t$ and $\mathbf{a}_t$ are the states and actions in the trajectory $H_k$ (we use bold notation for $\mathbf{a}_t$ to highlight that it is the joint action of the ego-agent and the teammates), $g_j(H_k)$ is the `alternative return' of the behavior policy $P_k$ for the function $g_j$ in the trajectory $H_k$, $P_i(\mathbf{a}_t|s_t)$ is the probability of policy $P_i$ to pick action $\mathbf{a}_t$ at state $s_t$, and $P_k(\mathbf{a}_t|s_t)$ is the probability of policy $P_k$ to pick action $\mathbf{a}_t$ at %
$s_t$. 

In our multiagent \textit{type-based} approach, IS is, however, suboptimal since it does not consider the types of the other agents and the related policies, which can provide very useful information for improving the estimate in terms of quality and sample efficiency. We observe that types are unknown, but they can be inferred from the dataset $D$ of trajectories if explicitly considered by the estimator. If we assume the teammates' action selection \textit{independent} from those of the controlled agent (i.e., the probability a teammate selects an action in a state does not depend on the behavior/new policy of the controlled agent) and \textit{stationary} between when we collect the dataset and the optimization phase, 
then we can rewrite the IS estimator in terms of all actions $\mathbf{a}_t$ as:
\begin{align}
     \hat{\rho}_{i,j,k}^{\texttt{IS}} &\defeq g_j(H_k) \prod_{(s_t,\mathbf{a}_t)\in H_k} \frac{P_i(\mathbf{a}_t|s_t)}{P_k(\mathbf{a}_t|s_t)} \label{eq:IS2}\\
     &= g_j(H_k)\prod_{(s_t,\mathbf{a}_t)\in H_k} \frac{P_i^{(\ego)}(a^{(\ego)}_t|s_t)P_i^{(\notego)}(\mathbf{a}^{(\notego)}_t|s_t)}{P_k^{(\ego)}(a^{(\ego)}_t|s_t)P_k^{(\notego)}(\mathbf{a}^{(\notego)}_t|s_t)} \label{eq:IS_Types1}\\
     &= g_j(H_k)\prod_{(s_t,\mathbf{a}_t)\in H_k} \frac{P_i^{(\ego)}(a^{(\ego)}_t|s_t)}{P_k^{(\ego)}(a^{(\ego)}_t|s_t)} \label{eq:IS_Type2}
\end{align}
where $P_i^{(\ego)}$ is the policy of the ego-agent and $P_i^{(\notego)}$ is the joint policy of the teammates. The equality between Eq. \eqref{eq:IS2} and Eq. \eqref{eq:IS_Types1} follows from independence. %
The equality between Eq. \eqref{eq:IS_Types1} and Eq. \eqref{eq:IS_Type2} stems from stationarity, having that: $P_i^{(\notego)}(\mathbf{a}^{(\notego)}_t|s_t) = P_k^{(\notego)}(\mathbf{a}^{(\notego)}_t|s_t)$. %
As shown in Eq. \eqref{eq:IS_Type2}, however, the teammate types and related policies cannot be naturally integrated into IS.

\subsection{Performance estimation with doubly-robust importance sampling}
To explicitly consider teammate types and environment transition function in the performance estimator, we use the \textit{Doubly-Robust Importance Sampling} (DR)  \citep{jiang_doubly_2016,thomas2016dataefficient,levine_offline_2020,HuangJ20}. It combines importance sampling and the direct method by introducing an estimate of the Q-function $\hat{Q}_j^{P_i}(s,a)$ inside the importance sampling formula, reducing the variance, usually very high in standard IS. DR has form:
    \begin{align}
    &\dr\defeq \sum_{(s_t,\mathbf{a}_t)\in H_k} \gamma^t \left( w_{\leq{t}} (g_j(s_t, \mathbf{a}_t) - \hat{Q}_j^{P_i}(s_t, a^{(\ego)}_t)) +w_{\leq t-1}\hat{V}_j^{P_i}(s_t)\right) \label{eq:DR2}
    \end{align}
where $w_{\leq t} = \prod_{\bar{t}\leq t} {{P_i}^{(\ego)}(a^{(\ego)}_{\bar{t}} |s_{\bar{t}})}/{P_k^{(\ego)}(a^{(\ego)}_{\bar{t}}|s_{\bar{t}})}$ (\emph{importance weight}), $\hat{Q}_j^{P_i}(s_t, a^{(\ego)}_t)$ is the estimated state-action value of policy ${P_i}$ for constraint $g_j$ in state $s_t$ and the ego-action $a^{(\ego)}_t$, and $\hat{V}^{P_i}(s_t)$ is the estimated state value of policy ${P_i}$ for constraint $g_j$ in state $s_t$.

This estimator is unbiased if either the behavior policy $P_k$ is known or the model of the environment is known (that is, the estimates $\hat{Q}_j^{P_i}(s_t, a^{(\ego)}_t)$ and $\hat{V}_j^{P_i}(s_t)$ are perfect) \citep{jiang_doubly_2016, thomas2016dataefficient}. In our case, the second condition is not satisfied because it would require perfect knowledge about opponents' types and environment transition model, but the first condition is satisfied since we assume to know the behavior policy $P_k$ by which the trajectory has been collected. The idea developed in this work is to explicitly introduce the types of the agents and the transition model of the environment in $\hat{Q}_j^{P_i}(s_t, a^{(\ego)}_t)$ and $\hat{V}_j^{{P_i}}(s_t)$, which, importantly, still leave the estimator unbiased (because it only changes the estimates).  Considering that $(\ego)$ is the ego-agent and $(\notego)$ are the teammates, \new{to estimate ${Q}_j^{P_i}(s_t, a^{(\ego)}_t)$ we can use the relation \citep{he_opponent_2016}}:
    \begin{align}\label{eq:Q_estimate}
    Q_j^{P_i}(s_t,a^{(\ego)}_t) \defeq \sum_{a_t^{(\notego)}}{P_i}^{(\notego)}(a_t^{(\notego)} | s_t) \sum_{s_{t+1}}&T(s_t, \bm{a}_t, s_{t+1}) \\&[\new{g_j(s_t, \bm{a}_t)}+ \gamma \ee_{{a}^{(\ego)}_{t+1}}[Q_j^{P_i}(s_{t+1}, a^{(\ego)}_{t+1})]. \nonumber
    \end{align} 
We can similarly derive the relation for $V_j^{P_i}(s)$, or define it as the expected value, over the actions $a^{(\ego)}_t$ taken under the current policy ${P_i}$, of $Q_j^{P_i}(s_t,a^{(\ego)}_t)$.

\subsubsection{Estimation of transition model and teammate types}
To compute Eq. \eqref{eq:Q_estimate}, we need to estimate the transition model $T$ and the teammate types $P^{(\notego)}$.
Both are obtained from the dataset of trajectories $D$ using a Maximum Likelihood Estimate (MLE) approach. In particular, the transition model probabilities are computed as
    \begin{align}
    \hat{T}(s_t,\bm{a}_t, s_{t+1})& \defeq\frac{|( s_t,\bm{a}_t, s_{t+1} )|_D}{\sum_{s\in S}|(s_t,\bm{a}_t, s )|_D}.\label{eq:T_est}
    \end{align}
where $|( s_t,\bm{a}_t, s_{t+1} )|_D$ is the number of times in dataset $D$ the ego-agent transitions from state $s_t$ to state $s_{t+1}$ when the joint action $\bm{a}_t$ is performed, and $\sum_{s\in S}|( s_t,\bm{a}_t, s )|_D$ is the total number of time in $D$ the ego-agent is in state $s_t$ and the joint action $\bm{a}_t$ has been performed. On the other hand, the policy type of each teammate $(\notego)$ is selected by maximizing its log-likelihood in $D$: 
    \begin{align}
    \hat{P}^{(\notego)} \defeq \argmax_{P \in P_\mathcal{T}} \sum_{(s_t, {a}^{(\notego)}_t) \in D} \log P(a^{(\notego)}_t|s_t).\label{eq:P_ego_est}
    \end{align}

\subsubsection{Dataset split}
To correctly obtain an unbiased estimate in Eq. \eqref{eq:DR2}, we need to split the dataset into two subsets, $D_\text{train}$ and $D_{\text{val}}$ \citep{jiang_doubly_2016}, %
one for estimating $T$, ${P_i}^{(\notego)}$, and $Q_j^{P_i}$(Eqs. \ref{eq:Q_estimate}, \ref{eq:T_est}, \ref{eq:P_ego_est}), and one for estimating $\dr$, given the other elements. In our experiments, we split the given dataset using a manually selected ratio $\lambda$ (usually between 0.15 and 0.55, see Section \ref{sec:results}).

\subsection{Lower-bound performance estimation via concentration inequalities}\label{sec:bounds}

To solve the Seldonian optimization problem of Eq \eqref{eq:Seldonian}, we must now find a way to guarantee the probabilistic constraints over \new{desirable} behaviors, namely, for each candidate policy $P_i$ we must check that $\forall j \in \{1, \ldots, n\} \ Pr(g_j(P)) \geq 0) \geq 1-\delta_j$. We do so by applying finite-sample concentration inequalities to get a \textit{lower bound} on the true estimated value $g_j(P_i)$ that holds probabilistically with the desired confidence level $\delta_j \in [0,1]$. We consider two concentration inequalities, whose formal definitions are reported below.

\paragraph{Extended Maurer \& Pontil's empirical Bernstein inequality \citep{ maurer2009bernstein, thomas_high-confidence_2015}:} 

\begin{align}\label{eq:bernstein}
{\rho}_{i,j}^{\texttt{DR}} \geq \left(\sum_{k=1}^m \frac{1}{\xi_k}\right)^{-1} \left[\sum_{k=1}^m \frac{Y_k}{\xi_k} - \frac{7m\log(2/\delta)}{3(m-1)} - \sqrt{\frac{2\log(2/\delta)}{m-1}\left(m\sum_{k=1}^m \left(\frac{Y_k}{\xi_k}\right)^2 - \left(\sum_{k=1}^m \frac{Y_k}{\xi_k}\right)^2\right)}\right]
\end{align}
where ${\rho}_{i,j}^{\texttt{DR}}$ is the true mean return, ${\rho}_{i,j}^{\texttt{DR}}=\ee[\dr] = g_j(P_i)$, $m$ is the number of trajectories over which ${\rho}_{i,j}^{\texttt{DR}}$ is computed (given the candidate policy $P_i$ and the undesired behavior $g_j$), $Y_k \defeq \min\{\dr, \xi_k\}$, where $\dr$ are our estimated returns for trajectory $k$ (see line 3 of Algorithm \ref{alg:our_seldonian}) and $\xi_k$ are real-value constants that must be tuned to achieve the tightest possible bound. We set this value by hand in all our experiments (see Section \ref{sec:results}). The lower bound we are looking for is the right-hand side of Eq. \eqref{eq:bernstein}, and this holds with probability $1-\delta_j$.

Eq. \eqref{eq:bernstein}, however, requires $\dr$ being unbiased and \textit{almost surely non-negative}, $\pp(\dr \geq 0) = 1$ (\cite{thomas_high-confidence_2015}, Theorem 1)\footnote{Importantly, it does not require the variables to be identically distributed, so it can be used with estimates obtained with different behavior policies.}. The first one is guaranteed by the properties of the DR estimator, the second one is not (due to the minus sign in Eq. \ref{eq:DR2} and the arbitrary values that the estimates can take). We can, however, add a constant $a$ to every estimated target $\dr, k=1, \ldots, m$ to force this property on the new estimator $\hat\psi_{i,j,k} = \dr+ a$. Of course, this makes it also biased, but by a known quantity $a$. The left-hand-side of Eq.  \eqref{eq:bernstein} becomes $ \ee[\hat\psi_{i,j,k}] = \ee[\dr+a] = \ee[\dr]+a = {\rho}_{i,j}^{\texttt{DR}} + a$ and we can remove the constant $a$ from the right-hand-side with a suitable redefinition of $Y_k$ to obtain a valid lower-bound for ${\rho}_{i,j}^{\texttt{DR}}$.
\paragraph{Lemma 1}
    If $\hat\psi_{i,j,k} = \dr + L(R^{\text{max}} + 2V^{\text{max}})$, with $L$ being the trajectory length, $R^{\text{max}}$ and $V^{\text{max}}$ respectively the maximum reward and value, then $\pp(\hat\psi_{i,j,k} \ge 0) = 1$.
    
\textit{Proof sketch.} The idea behind this proof is to identify a constant $a$ larger than or equal to $|\min \dr|$ and to use it to make all terms positive. By analyzing the terms in the sum of Eq. \ref{eq:DR2} it turns out that $|\min \dr| \leq L(R^{\text{max}} + 2V^{\text{max}})$. Full mathematical details are reported in Supp. Mat. Section \ref{sec:proof}.

\paragraph{Student's t-inequality \citep{student1908probable}:}
\begin{equation}\label{eq:t_student} 
    {\rho}_{i,j}^{\texttt{DR}} \geq \text{mean}(\boldsymbol{\hat\rho}) - \sqrt{\frac{\text{stdev}(\boldsymbol{\hat\rho})}{|\boldsymbol{\hat\rho}|}}\cdot \mathtt{T}^{-1}_{|\boldsymbol{\hat\rho}| - 1}[1 - \delta_j]
\end{equation}
where $\boldsymbol{\hat\rho}$ is the collection of estimates $\{\dr\}_{k=1}^n$ and $\mathtt{T}^{-1}_{d}[p]$ is the inverse CDF of the \textit{t-student} distribution with $d$ degrees of freedom for quantile $p$%
; \emph{mean} and \emph{stdev} are the empirical mean and standard deviation. This inequality, which holds with probability $1-\delta_j$, assumes that the provided estimates $\{\dr\}_{k=1}^n$ are distributed, in the limit, as a Gaussian, so it does not give the same formal finite-sample regime guarantees as Eq.~\eqref{eq:bernstein}; however, it is consistently used in many fields and also by \citep{thomas_preventing_2019}, so we consider it as a possible estimate in our work. In the following, we will call $\alpha_{i,j}$ the right-hand side of this inequality and that of Eq.~\eqref{eq:bernstein}.

\begin{algorithm}[ht]
\caption{Seldonian Ad Hoc Teamwork}\label{alg:our_seldonian}
\KwData{Dataset $D = \{(H_k, P_k )\}_{k=1}^m$, functions $\{g_j\}_{j=1}^n$, confidence levels $\{\delta_j\}_{j=1}^n$, candidate policies set $\{P_i\}_{i=1}^\ell$, possible teammate policies set $\{P_{\mathcal{T}_u}\}_{u=1}^q$, \new{number of teammates $p$}, split ratio $\lambda$.}
\KwResult{Either a reliable policy $P^\star$ or \textsc{No Solution}}
$\mathcal{P} \gets \emptyset$ (set of reliable policies)\\
Split ${D}$ into ${D}_{\text{train}}$ with ${D}_{\text{val}}$ using $\lambda$\\
Estimate transition model $T$ using Eq. \eqref{eq:T_est} on ${D}_{\text{train}}$\\
Estimate teammate types and related policies $P^{(\notego)}$ using Eq. \eqref{eq:P_ego_est} on ${D}_{\text{train}}$\\
\For{$i=1,\ldots, \ell$}{
    \For{$j=1,\ldots, n$}{
        \new{Estimate $Q_j^{P_i}(s_t,a^{(\ego)}_t)$ and $V_j^{P_i}(s_t)$ of candidate policy $P_i$ using Eq. \eqref{eq:Q_estimate} on ${D}_{\text{train}}$}\\
        Compute $m$ positive unbiased estimates $\{ \dr \}_{k=1}^m$ of $g_j({P_i})$ using Eq. \eqref{eq:DR2} + Lemma 1 on ${D}_{\text{val}}$\\
        Compute the lower bound $\alpha_{i,j}$ of $g_j(P_i)$ with confidence level $\delta_j / \ell$  using Eqs. \eqref{eq:bernstein} or \eqref{eq:t_student} (it holds with confidence level $\delta_j$ simultaneously for all $\ell$ candidates \cite{thomas_preventing_2019}) \\
    }
    \If{$\alpha_{i,j} > 0,  \forall j=1, \ldots, n$}{Put $P_i$ into $\mathcal{P}$ }
}
\eIf{$\mathcal{P}$ is not $\emptyset$}{
    \For{$P \in \mathcal{P}$}{
        Estimate expected return performance $r(P)$ of $P$ using either IS or DR%
    }
    \Return $P^\star \defeq \argmax_{P\in \mathcal{P}} r(P)$
}
{
    \Return \textsc{No Solution}
}
\end{algorithm}

\subsection{Seldonian Ad Hoc Teamwork algorithm}

The proposed reliable Ad Hoc Teamwork approach, inspired by \citep{thomas_preventing_2019}, is illustrated in Algorithm \ref{alg:our_seldonian}. The algorithm receives in input a dataset of trajectories $D = \{(H_k, P_k )\}_{k=1}^m$, a set of functions $\{g_j\}_{j=1}^n$ that evaluate \new{desirable} behaviors, a set of associated confidence levels $\{\delta_j\}_{j=1}^n$, a set of candidate policies $\{P_i\}_{i=1}^\ell$, a set of possible teammate policies $\{P_{\mathcal{T}_u}\}_{u=1}^q$, \new{the number of teammates $p$}, and the split ratio $\lambda$. The output is the candidate policy $P$ having the highest estimated return among those that satisfy the probabilistic constraints $Pr(g_j(P) \geq 0) \geq 1-\delta_j$, or \textsc{No Solution} if no policy satisfies the constraints. The algorithm is divided into two parts: 
\begin{enumerate}
    \item from line 5 to 13, it computes the set of reliable policies, that is, those that satisfy the constraint. To do so, it computes, in {line 8}, {$m$ unbiased estimates of $g_j(P_i)$} using the $m$ trajectories in the dataset $D$, using the DR estimator Eq. \eqref{eq:bernstein}. This is done for each candidate $P_i$, and considering all the possible types of the teammates; then, in line 9, it uses the $m$ estimates to compute a {lower bound $\alpha_{i,j}$} on the \textit{true expected value} using a finite-sample concentration inequality (see Section \ref{sec:bounds}) for each constraint. Lastly, if the lower bound satisfies all constraints (line 11), the candidate is put into the set of \textit{reliable} policies.
    \item from line 15 to 22, if the set of reliable policies is not empty, then a similar procedure is carried out: we use the dataset $D$ to compute the estimated return $r(P)$ of each of the found reliable policies (using IS or DR), and we return the best one; otherwise, we return \textsc{No Solution} (lines 19 and 21).
\end{enumerate}

\section{Experimental Evaluation}\label{sec:results}

In this section, we present the results of our tests on three different settings, of increasing level of complexity: \textit{Chain World} \citep{chalkiadakis_coordination_2003}, \textit{Blackjack} \citep{sutton2020rl, towers2024gymnasium}, and \textit{Level-Based Foraging} \citep{papoudakis2021benchmarking, christianos2020shared}.

We compare our method against two baselines: an unreliable estimator that skips the reliability computation and picks the estimated best policy, similarly to the baseline used by \cite{thomas_preventing_2019}, and Algorithm \ref{alg:our_seldonian} using Per-Decision Importance Sampling estimator (PDIS) estimate instead of DR. PDIS \citep{doina2000Eligibility} is more efficient than standard IS, and similarly unbiased. It works by computing the importance weights incrementally (more details in Supp. Mat. \ref{sec:PDIS}). However, this estimator does not allow us to explicitly represent the transition model and the teammate types, thus, it is less efficient than our method. We chose PDIS instead of IS to make the comparison more fair, considering the poor results of IS in our tested environments in early tests. Since the Seldonian RL algorithm proposed by \cite{thomas_preventing_2019} (Fig. S20) uses a (modified) version of IS, and does not take into account AHT knowledge, we do not use it as a baseline in our experiments.

\subsection{Chain World}\label{exp:chain_world}
\begin{figure}[t]
  \centering
  \includegraphics[width=0.45\linewidth, trim={0 -0.5cm -.5cm 0}]{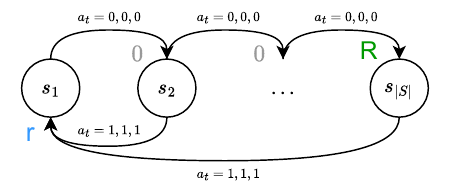} 
  \includegraphics[width=0.265\linewidth, trim={0 0 0 0}]{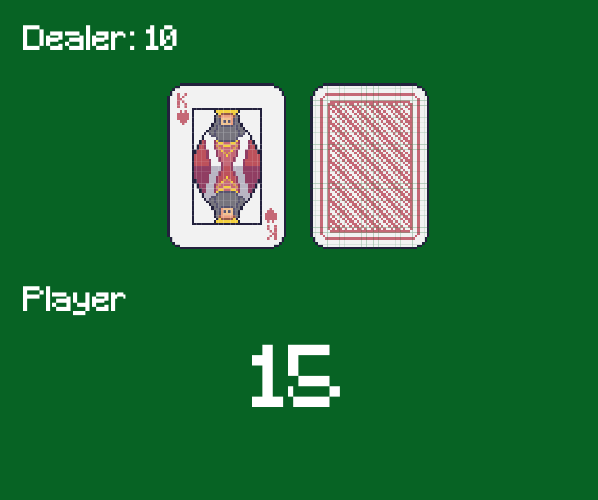} 
  \includegraphics[width=0.262\linewidth,  trim={-2cm 0 0 0}]{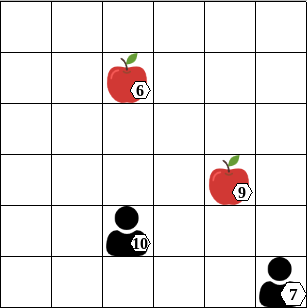} 
  \caption{Environments tested: \textit{Chain World}, a modified two-player version of \textit{Blackjack} and \textit{Level-Based Foraging}.}
  \label{fig:chain_world}
\end{figure}

{\em Chain World} \citep{chalkiadakis_coordination_2003} (Figure \ref{fig:chain_world}, left) is a coordination game in which there are $|S|$ states, $k$ agents and two actions, $0$ and $1$. Players start at state $s_1$. 
If all agents pick action $0$ in state $s_i$, they move to the next state $s_{i+1}$ and obtain zero reward. If they all pick action $1$, they go back to the starting state $s_1$ and receive a small reward $r$. If they pick conflicting actions, they stay in the same state and get zero reward. The goal is to coordinate completely by picking action $0$ $|S| - 1$ times and reach the end state, where they get a very large reward $R$. 

In our experiments, we set $|S| = 10, k=3, r=10$ and $R=100$. We also define a single function $g$ which measures the level of non-coordination between the agents. 
In \textit{Chain World}, we reward agreement among agents by defining $g(s_t, \bm{a}_t) = \exp(-x)$, where $x = 0$ if all agents' actions are identical, and $x=1$ otherwise. This creates two contrastive goals: maximizing the reward does not imply maximizing the agreement; if the teammates, for example, are more likely to take action $0$ instead of $1$, then an ego-policy that is more likely to take $1$ can be better in terms of reward but worse in terms of agreement.
We use fixed-length episodes ($L = 200$) by having the agents restart the game to $s_1$ when reaching the last state and add stochasticity to the transitions.

\textbf{Results.}\label{exp:results:chain_world} 
For this environment, we handcraft five rule-based policies $P_1,\dotsc, P_5$. These policies differ in their likelihood of picking different actions depending on the state $s_i$: some have a fixed probability, and others have a probability that depends on the state $s_i$. We pick two of these policies as the actual policies of the teammates, and we do not change them between the offline collection phase and the inference phase. 
Moreover, instead of requiring $g \ge 0$ as the probabilistic constraint in the second part of Eq \eqref{eq:Seldonian}, we consider a bound of the form $g \ge d $, where $d$ is an arbitrarily-picked constant chosen so that only a subset of the five policies considered as candidates are reliable.
We collect data using three different behavior policies, randomly taken from our set of five handcrafted policies. In particular, for each behavior policy, we collect 20 datasets for each of the five different sizes $|s| \in \{20,200,500, 1000,2000\}$. As for the candidate policies, we use the remaining four policies (besides the behavior). We pick all five policies as possible types from which we need to estimate the true teammate types. 
We test two inequalities for the lower bound: empirical Bernstein and Student's t-inequality (Eqs. \ref{eq:bernstein}, \ref{eq:t_student}). We pick $\delta = 0.15$ for both and set the split ratio $\lambda = 0.15$.

\begin{figure}[t]
    \centering
    \includegraphics[width=0.99\textwidth]{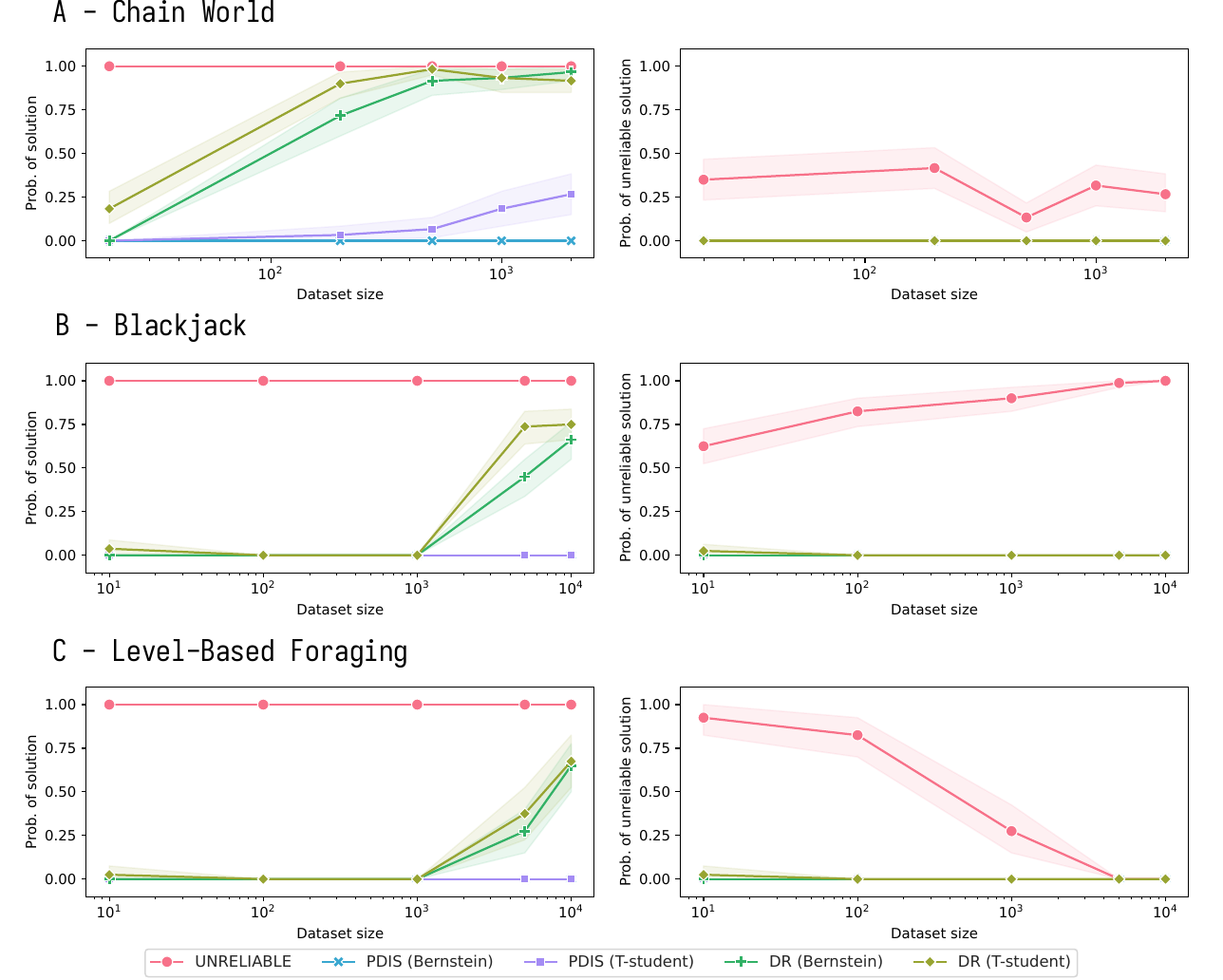}

    \caption{Results on the three tested environments. Different colors correspond to different \textit{estimators} used and different \textit{concentration inequalities} for the lower bound.
    On the x-axis, the size of the dataset. On the y-axis, on the left, the probability of returning a candidate policy as a solution --- higher is better; on the right, the probability of returning a \textit{unreliable} policy --- lower is better. The shaded areas represent the 95\% confidence intervals over multiple runs.}
    \label{fig:results_chainworld}
\end{figure}

We present the results of our method (labeled DR) in Figure \ref{fig:results_chainworld}, similar to \cite{thomas_preventing_2019}. Our algorithm (dark and light green curves, plot \textbf{A}) is consistently reliable. In fact, the plot on the right shows it never picks an unreliable policy, even with small dataset sizes, whereas the unreliable baseline consistently picks bad policies, with a probability between 10\% and 25\%, depending on the dataset size $s$. Notice that the PDIS-based method also reaches full reliability, but due to trivially rarely picking a policy (left plot, blue and purple curves). The DR estimator is indeed much more efficient than PDIS, with a higher probability of picking a solution for all $|s|$ (x-axis). Moreover, the two charts show that, by using the Student's t-inequality to compute the performance lower bound, our algorithm needs less data to be able to pick policies, reaching, for instance, a prob. of solution of around 25\% with a dataset size of only 20 episodes, while still avoiding unreliable policies. Because the unreliable baseline skips the reliability computation, it picks solutions with probability 100\%.

\subsection{Blackjack} \label{sec:blackjack}
Next, we test the scalability of our algorithm and its ability to work with trained policies. We do it by applying the algorithm to a more complex environment, a modified multiplayer version of Blackjack \citep{sutton2020rl, towers2024gymnasium} (Figure \ref{fig:chain_world}, center).
Blackjack is a stochastic environment where a single agent plays against the dealer, with the goal of drawing a hand of greater value without going \textit{bust} (over 21 points).
In our modified version, we add a second teammate who picks among the two actions (\textit{hit} or \textit{stick}) independently of the ego-agent. If both pick the same action, then the game proceeds as usual; otherwise, a turn passes without anyone drawing any card. In that case, the ego-agent and the teammate receive a small reward $r$ (globally). The game ends if either the players go bust (which makes them receive a 0 reward, globally) or if the dealer goes bust (in this case, they receive a large reward $R$). Alternatively, the game ends when $L$ turns pass without anyone going bust; in this case, the hands' values are compared to determine the winner. Having to coordinate with a second player can change the ego-agent strategy from the standard one: for example, there might be cases where, even if it would prefer to take the \textit{stick} action, knowing that the teammate is likely to take the \textit{hit} action could lead him to decide to coordinate with him instead to avoid moving to the next turn with no cards drawn, to avoid the end of the episode.

We set $L=10, r=0.5$ and $R=5$. We also make the environment fully observable by showing all the cards at all times. In terms of dimension, this environment has $8192$ state-action pairs, a much greater number than our chain world environment.
As for $g$, we define a single alternative 
return
function $g$ s.t. $g(s_t, \mathbf{a}_t) = 1$ if both agents pick the same action, 0 otherwise, rewarding agreement.

\textbf{Results.}\label{exp:results:blackjack}
Following the same experimental setting of the previous section, we handcraft three rule-based stochastic policies that pick actions based on the player's hand value and use them as possible types for the teammate (we pick one as the actual followed policy and keep it fixed during all phases).
In terms of ego-agent candidate policies, instead of only testing handcrafted policies, we also test trained deep RL policies. In particular, we train a PPO policy \citep{schulman2017ppo} over 1M steps, take two checkpoints at different levels of training (medium and high), and create eight different $\varepsilon$-greedy policies by adding variability in their action selection. We use four of these policies as candidates and four as behaviors. As in our previous experiment, we define reliability as achieving an expected $g \ge d$, where $d$ is handpicked to let only one policy among the candidates be reliable.
For this environment, we choose dataset sizes $|s| \in \{10, 100, 1000, 5000, 10000\}$, split ratio $\lambda = 0.55$ and confidence level $\delta = 0.05$. Given the large state-action space, we follow the standard procedure \citep{jiang_doubly_2016} to reduce the variance of all estimators by limiting the range of possible estimates between $[V_\text{m}, V_{M}]$ using domain knowledge. We run our algorithm 20 times for each size $|s|$ and each behavior, obtaining 95\% confidence intervals (represented by the shaded areas).

The results can be seen in Figure \ref{fig:results_chainworld}, plot \textbf{B}. We see, similarly to our previous experiment, that the unreliable baseline always selects a policy (left plot) but it often picks unreliable policies (right plot) because it focuses only on the reward, a goal not always aligned with the player's agreement. Our algorithm (green lines) requires more data (i.e., $|s| \ge 5000$) to start picking policies (left plot), but when it does, it is consistently reliable (right plot). Only when the dataset is very small and we use Student's t-inequality our algorithm can pick, still rarely, bad policies (right plot). This does not happen when we use the extended Bernstein inequality (Eq. \ref{eq:bernstein}). In general, the latter is more conservative and we can clearly see that, by using it, our algorithm is less likely to pick a policy than when we use the looser t-student (i.e., the dark-green line is lower than the light-green one in the left plot). Still, when $|s| = 10000$, our algorithm picks a reliable policy with a 100\% chance (or it selects no policy). Clearly, the PDIS estimator is not good enough at this size as it cannot find good enough estimates to even pick a single policy in all experiments and for all concentration inequalities.

\subsection{Level-Based Foraging} \label{sec:lbf}
As a third experiment, we scale even further by testing our algorithm on \textit{level-based foraging} (Figure \ref{fig:chain_world}, right), a standard AHT benchmark \citep{papoudakis2021benchmarking, christianos2020shared}. This environment is much bigger than the others, with more than $31k$ state-action pairs. In level-based foraging, $n$ players must collaborate to collect $m$ foods. Players and foods are assigned different levels, and one or more players can collect a piece of food only by getting close to it and selecting the \textit{load} action. When this happens, the total sum of the players' levels $S$ is compared with the food level $f$: they succeed only if $S \ge f$, at which point they receive reward $R = f$. Every other action obtains zero reward.
In our experiments, we pick $n=2$ and $m = 2$. We also make the environment of fixed length, setting $L=25$. As a desired constraint function, we use the expected return, that is, the reliability is based on the reward itself. This definition is allowed by Eq. \eqref{eq:Seldonian} and lets the users of our algorithm capture the cases where they want guaranteed lower bounds on the reward before committing to a policy. This form was already explored, albeit slightly differently, in \citep{thomas_preventing_2019}. As before, we require a policy to have $g \ge d$, for a handpicked $d$, to be reliable.

\textbf{Results.} Again, we follow our previous experimental setting. We consider three handcrafted policies for the opponent types, one of which is randomly selected as the actual policy. For the candidates and behaviors, we train PPO \citep{schulman2017ppo} for 20M steps, pick three different checkpoints, and create seven different $\varepsilon-$greedy policies out of them. We pick four as candidates and three as behaviors. We also pick the same set sizes, split ratio, and confidence level, and we also limit the estimates for each estimator between $[V_{m}, V_{M}]$ \citep{jiang_doubly_2016}. We run our algorithm 10 times each for each size and behavior, obtaining 95\% CI (shaded areas).

The results can be seen in Figure \ref{fig:chain_world}, plot \textbf{C}. Owing to the more difficult estimation target, due to the environment size, we see that our algorithm needs a large enough dataset size $|s| \ge 5000$ to consistently pick a reliable policy (left plot). Using Student's t-inequality makes our algorithm less conservative and reaches around 75\% probability of picking reliable policies with $|s| = 10000$. If we use the extended Bernstein inequality, we get similar results but slightly worse, especially when $|s| = 5000$. This is expected as it is known to require more data \citep{thomas_preventing_2019, thomas2016dataefficient} than alternatives with fewer guarantees (like Student's t-inequality). In general, it is standard to use a looser bound when reliability is important but not mandatory, and the environment is so big that collecting large enough datasets is prohibitive; in particular, the Student's t-inequality is used even in hard scientific field \citep{thomas_preventing_2019}. The unreliable has 100\% probability of picking policies, but only when the dataset size $|s|$ gets big ($|s| \ge 5000)$ then its probability of picking unreliable policies goes to zero. This happens because, by defining the constraint as the reward itself, the unreliable baseline starts correctly estimating the policy performance as the dataset size gets larger, and hence, by picking the candidate with the largest estimated return, it correctly picks reliable policies. Our algorithm, instead, never picks bad policies so it is reliable in this setting.

\section{Related Work}\label{sec:related}

\paragraph{Seldonian optimization} The original Seldonian optimization framework was proposed by \cite{thomas_preventing_2019}. %
After proposing a shift in perspective in ML algorithmic design, from performance optimization to avoiding undesirable behavior, the authors show how to apply this principle to build, in a proof-of-concept diabetes management simulated environment, a safe RL algorithm that avoids suggesting policies leading to dangerous low blood sugar levels. Other, more recent works have focused on extending this approach to handle more specific RL settings. In \cite{satija2021multiobjectivespibb}, the authors implement an offline Seldonian algorithm in the Safe Policy Improvement setting \citep{thomas_hcpi_2015, ghavamzadeh2016spi, laroche2019spibb, castellini2023spi, bianchi2024spi,Bianchi_AIRO_2024}, by casting the task as a multiple objective optimization problem in a Constrained MDP \citep{altman1999CMDP}. Their algorithm can return, with statistical guarantees, a new policy that performs at least as well as the baseline policy used to collect the data, for any considered objective. The work by \cite{chandak2020nonstationary}, instead, extends the Seldonian problem to handle non-stationary MDPs, using a candidate policy search plus a safety test procedure. None of these works consider the multiagent \citep{marl-book} or the AHT setting \citep{stone2010adhoc}.
\paragraph{Ad Hoc Teammwork} Most of the algorithms applied to the AHT problem \citep{stone2010adhoc, albrecht_autonomous_2018, mirsky_survey_2022} are not able to provide reliability guarantees. State-of-the-art algorithms \new{for multiagent domains, which can be used to solve AHT problems,} include SEAC  \citep{christianos2020shared}, MAPPO \citep{Chao2021MAPPO} and MAA2C \citep{papoudakis2021benchmarking}. In terms of \textit{type-based} AHT \citep{albrecht_autonomous_2018}, which is the approach that we take, related works usually adopt a \textit{Bayesian} point-of-view, by considering beliefs on the types of the teammates, updated with each subsequent observation \citep{chalkiadakis_coordination_2003, albrecht2019reasoning} or implicitly model the teammates using neural-network approximations \citep{he_opponent_2016}. More recently, some works for the AHT have explored providing guarantees, for example, \textit{robustness} and \textit{worst-case performance} \citep{rahman2024coverageset, villin2025minimax}. However, none of these works tackle the \textit{offline} setting and the Seldonian optimization framework.

\section{Conclusions and future work} \label{sec:conclusion}
We presented a novel offline RL algorithm that, given enough data, can return a reliable policy with respect to predefined \new{desirable} behaviors in non-coordinated environments. The approach solves a Seldonian optimization problem in the context of Ad Hoc Teamwork. We showed experimentally that the technique can scale to more complex environments and deal with any type of candidate policy. Still, our approach presents some limitations that stimulate interesting future extensions. %

First, in its current form, the proposed algorithm needs both a set of \textit{candidate} policies for the ego-agent and a set of \textit{possible types} for the teammates. 
Such assumptions can sometimes be unrealistic and could place an unnecessary burden on the user of the algorithm. However, domain knowledge is often available and can allow for informed priors that are key for identifying possible candidate policies \citep{Bonanni_AAMAS_2025}. Also, while knowing the possible teammates' types helps with sample efficiency, it is not crucial for the algorithm. In any case, we intend to
explore different ways to relax such requirements, such as searching directly in the space of neural network policies for the candidates or estimating the types of teammates using an MLE approach. Second, solving for the exact Q-value function in Eq \eqref{eq:Q_estimate} can be a bottleneck, in terms of scalability and computational requirements. In our experiments, we tried different approximators, such as neural networks and XGBoost regressors \citep{cen2016xgboost}, with negative outcomes. We believe that investigating these approaches further can be key to scaling the algorithm even more, with the goal of reaching real-world scenarios. Finally, %
extending our approach to the fully online setting is interesting future work.

\section*{Acknowledgments}
The research described in this paper was carried out within the framework of the National Recovery and Resilience Plan Greece 2.0, funded by the European Union - NextGenerationEU (Implementation Body: HFRI. Project name: DEEP-REBAYES. HFRI Project Number 15430).

We also acknowledge financial support from PNRR MUR project PE0000013-FAIR and PR Veneto FESR 2021-2027 24729-002238 - SINERGHY.

\appendix

\bibliography{biblio}
\bibliographystyle{rlj}

\beginSupplementaryMaterials
\section{Per-Decision Importance Sampling}\label{sec:PDIS}
The Per-Decision Importance Sampling estimator (PDIS) is a similar estimate to IS, unbiased under the same assumptions, but with lower variance. Its formula is:
\begin{equation}
\begin{aligned}
    \new{{\hat\rho}_{i,j,k}^{\texttt{PDIS}} = \sum_{(s_t,\mathbf{a}_t,r_t) \in D} \gamma^t g_j(s_t, \mathbf{a}_t) w_{\leq t}}
\end{aligned}
\label{eq:pdis}\end{equation}
where $w_{\leq t} \defeq \prod_{\bar{t}\leq t} {P_i(\mathbf{a}_{\bar{t}} |s_{\bar{t}})}/{P_k(\mathbf{a}_{\bar{t}}|s_{\bar{t}})}$. The difference with IS is that the importance weight is not calculated up to the end of the trajectory in one shot, but incrementally as a function of the steps up to that point.

\section{Proof of Lemma 1}\label{sec:proof}

\paragraph{Lemma}
    If $\hat\psi_{i,j,k} = \dr + L(R^{\text{max}} + 2V^{\text{max}})$, with $L$ being the trajectory length, $R^{\text{max}}$ and $V^{\text{max}}$ respectively the maximum reward and value, then $\pp(\hat\psi_{i,j,k} \ge 0) = 1$.

\textit{Proof} 
For any random variable $X$, we have that $\pp(X + a \ge 0) = 1$ if $a \ge |\min X|$. Therefore, if we find a constant $a$ such that $a \ge |\min \dr |$ then $\pp(\dr + a \ge 0) = 1$. First, for simplicity we rewrite $\dr$ from Eq. \eqref{eq:DR2} as:
\begin{align}
    \dr &=\sum_{(s_t,\mathbf{a}_t)\in H_k} \left(\textcolor{black}{\gamma^t w_{\leq{t}} g_j(s_t, \mathbf{a}_t)} - \textcolor{black}{\gamma^t w_{\leq{t}} \hat{Q}_j^{P_i}(s_t, a^{(\ego)}_t)} + \textcolor{black}{\gamma^t w_{\leq t-1}^{H_k}\hat{V}_j^{{P_i}}(s_t)} \right)\nonumber \\ 
    &= \sum_{(s_t,\mathbf{a}_t)\in H_k} (\textcolor{black}{X_t} - \textcolor{black}{Y_t} + \textcolor{black}{Z_t})
\end{align} 
We have that:
\begin{align}
    |\min \dr| &\le \max| \dr|\\
    &= \max\left|\sum_{(s_t,\mathbf{a}_t)\in H_k} (\textcolor{black}{X_t} - \textcolor{black}{Y_t} + \textcolor{black}{Z_t})\right|\\
    &\le \max\sum_{(s_t,\mathbf{a}_t)\in H_k}\left| \textcolor{black}{X_t} - \textcolor{black}{Y_t} + \textcolor{black}{Z_t}\right|\label{eq:firstsumexplanation}\\ 
    &\le \max\sum_{(s_t,\mathbf{a}_t)\in H_k} (\textcolor{black}{X_t} + \textcolor{black}{Y_t} + \textcolor{black}{Z_t}) \label{eq:secondsumexplanation}\\
    &\le \sum_{(s_t,\mathbf{a}_t)\in H_k} \max(\textcolor{black}{X_t} + \textcolor{black}{Y_t} +\textcolor{black}{Z_t}) \label{eq:maxexplanation}\\
    &\le \sum_{(s_t,\mathbf{a}_t)\in H_k} (\max\textcolor{black}{X_t} + \max\textcolor{black}{Y_t} + \max\textcolor{black}{Z_t}) \label{eq:secondmaxexplanation}\\
    &\le \sum_{(s_t,\mathbf{a}_t)\in H_k} (R^{\text{max}} + V^{\text{max}} + V^{\text{max}}) \label{eq:finalstepexplanation}\\
    &=L(R^{\text{max}} + 2V^{\text{max}})
\end{align}

Eq. \eqref{eq:firstsumexplanation} follows from $|\sum_i \alpha_i| \le \sum_i |\alpha_i|$. Likewise, Eq. \eqref{eq:secondsumexplanation} follows from $|a+(-b)+c| \le |a| + |-b| +|c| = |a|+|b|+|c|$ and the fact that $X_t, Y_t, Z_t$ are all positive, under the design choice of having only positive rewards (because all its terms are positive, although we require to constraint the \textit{estimated} Q-value and Value functions to be positive, which is easily done).  Eqs. \eqref{eq:maxexplanation} and \eqref{eq:secondmaxexplanation} follow from $\max(\sum_i \alpha_i)\le \sum_i\max \alpha_i$. 

Lastly, Eq. \eqref{eq:finalstepexplanation} follows from the fact that $\gamma^t, w_{\leq{t}}, w_{\leq{t-1}} \in [0,1]$ for all $t$ and all trajectories, and that $g_j(s_t, \mathbf{a}_t) \le R^{\text{max}}$ and $\hat{Q}_j^{P_i}(s_t, a^{(\ego)}_t),\hat{V}_j^{P_i}(s_t) \le V^{\text{max}}$ under the very natural assumption that we constraint these \textit{estimated} functions to be below $V^{\max}$: this is trivially done, as we can just clip them --- this does not affect the properties of the DR estimator.
    
Therefore, if we set $a = L(R^{\text{max}} + 2V^{\text{max}})$ and  $\hat\psi_{i,j,k} =\dr + L(R^{\text{max}} + 2V^{\text{max}})$, the statement follows. $\square$
\section{Ablation study on the randomness of the behavior policy}
\begin{figure}[t]
    \centering
    \includegraphics[width=1\linewidth]{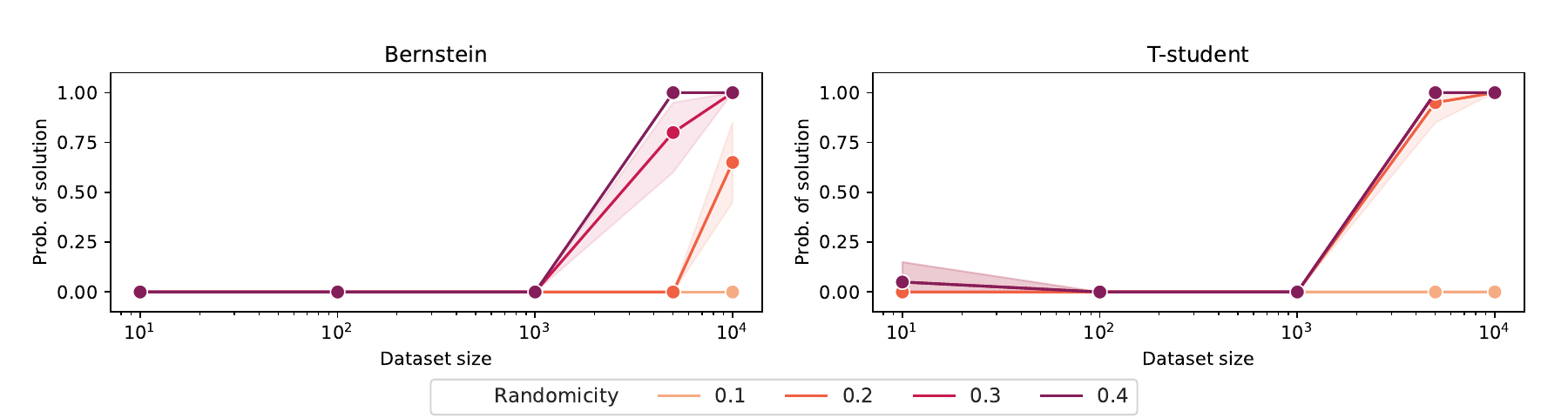}
    \includegraphics[width=1\linewidth]{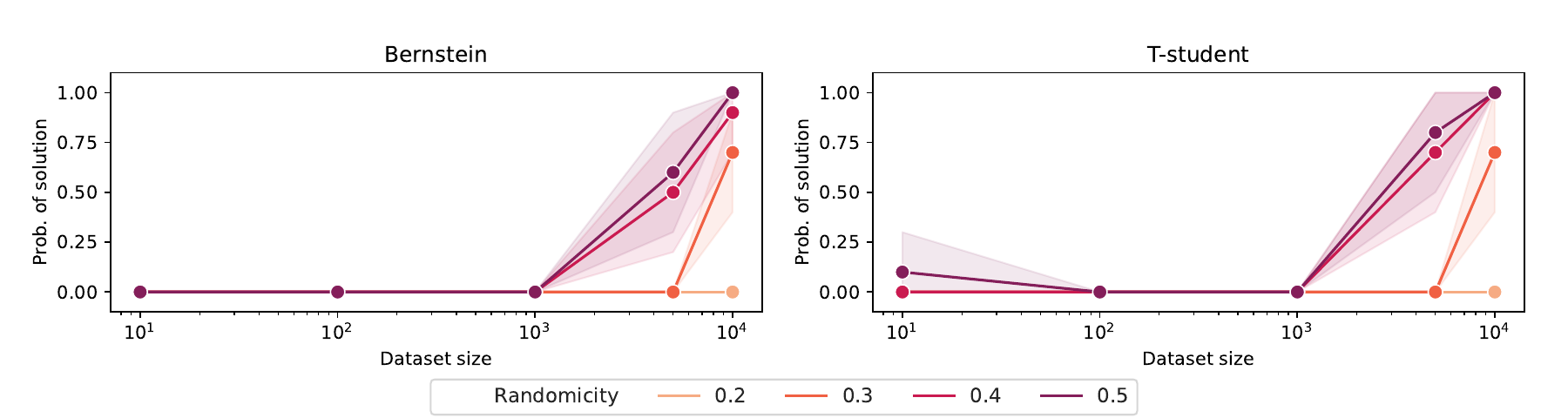}

    \caption{Probability of solution as a function of the randomness of the behavior policy. Top: \textit{Blackjack}, bottom: \textit{Level-Based Foraging}. Left: DR using Bernstein, right: DR using Student's  t-inequality.}
    \label{fig:results_randomicity}
\end{figure}
We found that the performance of our algorithm, especially the probability of solution, is sensitive to the randomicity of the behavior policy used to collect data; by randomicity, we mean the $\varepsilon$ stochasticity that we add to the raw NN-based policy (see Sections \ref{sec:blackjack} and \ref{sec:lbf}). Figure \ref{fig:results_randomicity} presents the probability of solution for our algorithm (using DR exclusively) based on this stochasticity. The plots are obtained by separating, using the same data of Figure \ref{fig:results_chainworld}, the results obtained using different behavior policies (similarly to Figure \ref{fig:results_extended}). That is, each line in Figure \ref{fig:results_randomicity} corresponds to $n$ runs of Algorithm \ref{alg:our_seldonian} (for \textit{Chainworld} and \textit{Blackjack}, $n = 20$, for \textit{Level-Based Foraging} $n=10$) \textit{using data collected by a single behavior policy}.

We can see that, for all inequality bounds (left, Bernstein, right, Student's t-inequality) and all environments (top, \textit{Blackjack}, bottom, \textit{Level-Based Foraging}), collecting data with a more random behavior policy leads to improved probability of solution. For both environments, values of $\varepsilon$ lower than 0.3 (light orange) lead to a very low probability of solution (0 in some cases), whereas by increasing the randomicity we get higher and higher values, up to 1.0 when the dataset size gets larger. The effect is clearer when we use the {Bernstein} inequality on \textit{Blackjack} (top-left plot). Still, this effect is present even when we use Student's t-inequality (right plots), even though sometimes the differences between different randomicity are not as clear (for example, in \textit{Blackjack}, top-right). 

This behavior can generally be expected as the higher the randomicity, the larger the state-action space explored, leading to more varied data for the out-of-policy estimations. 

\section{Extended results}
\begin{figure}
    \centering
    \includegraphics[width=1\textwidth]{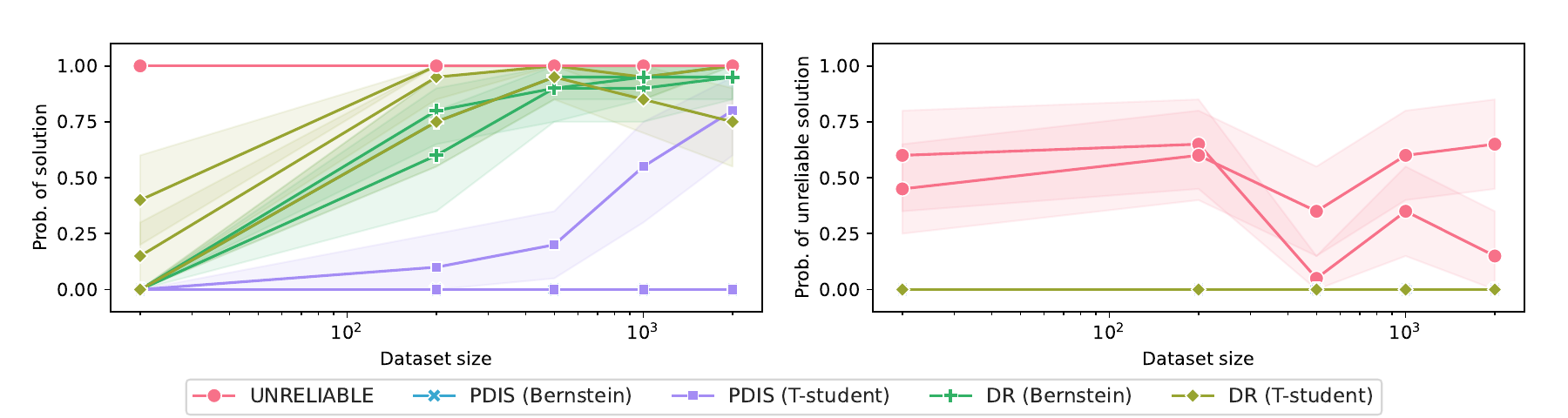}
    \includegraphics[width=1\linewidth]{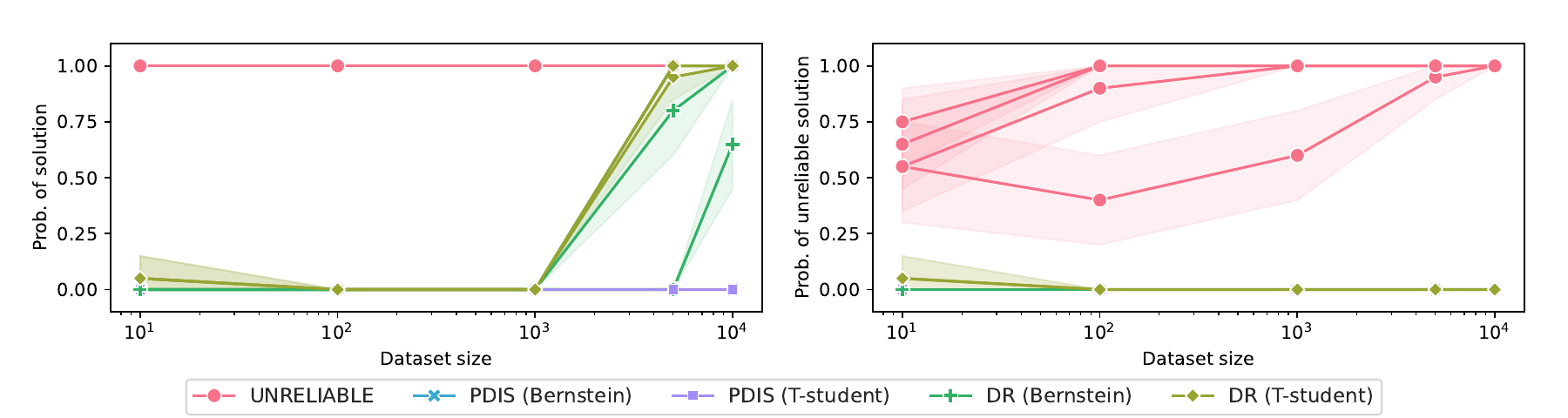}
    \includegraphics[width=1\linewidth]{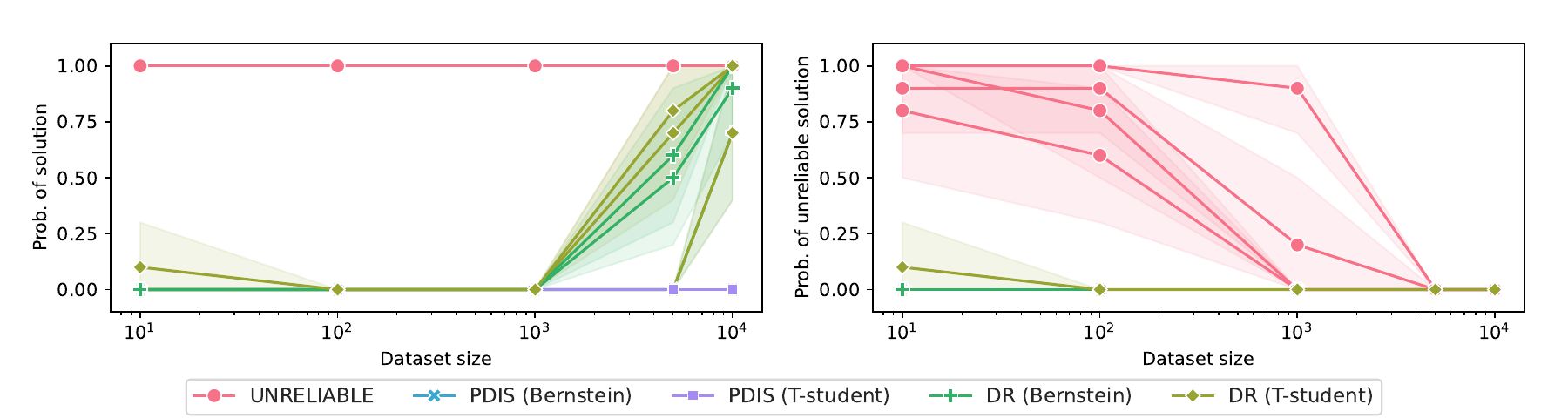}
\caption{Extended results for \textit{Chain World} (top), \textit{Blackjack} (center) and \textit{Level-Based Foraging} (bottom). Each line corresponds to $n$ runs of Algorithm \ref{alg:our_seldonian} using a different behavior policy. See text for more details.}\label{fig:results_extended}\end{figure}

In Figure \ref{fig:results_extended} we present the same results as in Figure \ref{fig:results_chainworld}, but we do keep data separated depending on the behavior policy that was used to collected it. That is each line in Figure \ref{fig:results_extended} corresponds to $n$ runs of Algorithm \ref{alg:our_seldonian}, where $n$ depends on the environment (for \textit{Chain World} and \textit{Blackjack} we have $n = 20$, for \textit{Level-Based Foraging} $n=10$) starting from data collected by a single behavior policy. The shaded areas represent the 95\% CI over the $n$ runs. This shows, alongside Figure \ref{fig:results_randomicity} (which presents explicitly the same curves for DR of the left plots --- the probability of solution --- based on the randomicity $\varepsilon$ of the behavior policies), that the results depend a lot on the behavior used to collect data. In some cases, using {Bernstein} is better than {Student's t-inequality} (bottom-left plot, which corresponds to the probability of solution for \textit{Level-Based Foraging}), but in general the latter is more efficient and lets us the Algorithm pick more policies and with less data. These plots also show that, across all domains and all runs, only a single time our Algorithm with PDIS, the second baseline, was able to get a probability of solution greater than zero, in \textit{Chain World} by using {Student's t-inequality} (top-left plot, purple line); in all other cases its probability of solution is zero. Meanwhile, by using {DR} we get much better estimations and we are able to pick solutions using almost all behaviors.
\end{document}

%% file: packages.tex
\usepackage[ruled,linesnumbered]{algorithm2e}
\SetKwComment{Comment}{/* }{ */}

\definecolor{phtalogreen}{HTML}{123524}
\definecolor{fgreen}{HTML}{228b22}
\definecolor{seablue}{HTML}{006994}
\definecolor{draftblue}{HTML}{116994}
\definecolor{algogreen}{HTML}{006400}
\definecolor{purpleblue}{HTML}{004194}
\definecolor{truepurple}{HTML}{800020}
\definecolor{mypurple}{HTML}{800020}
\definecolor{lightred}{HTML}{F77189}
\definecolor{lightblue}{HTML}{39A7D0}
\definecolor{lightgreen}{HTML}{36ADA4}

\newcommand{\new}[1]{\textcolor{black}{#1}}

\newcommand{\ego}{\text{ego}}
\newcommand{\notego}{-\text{ego}}